\documentclass[letterpaper]{article} % DO NOT CHANGE THIS
\usepackage{aaai2026}  % DO NOT CHANGE THIS
\usepackage{times}  % DO NOT CHANGE THIS
\usepackage{helvet}  % DO NOT CHANGE THIS
\usepackage{courier}  % DO NOT CHANGE THIS
\usepackage[hyphens]{url}  % DO NOT CHANGE THIS
\usepackage{graphicx} % DO NOT CHANGE THIS
\usepackage{natbib}  % DO NOT CHANGE THIS AND DO NOT ADD ANY OPTIONS TO IT
\usepackage{caption} % DO NOT CHANGE THIS AND DO NOT ADD ANY OPTIONS TO IT
\usepackage{algorithm}
\usepackage{algorithmic}
\usepackage{amsmath} 
\usepackage{booktabs}
\usepackage{multirow}
\usepackage{tabularx}
\usepackage{graphicx}
\usepackage{subcaption}
\usepackage{textcomp}
\usepackage{newfloat}
\usepackage{listings}
\DeclareCaptionStyle{ruled}{labelfont=normalfont,labelsep=colon,strut=off} % DO NOT CHANGE THIS
\floatstyle{ruled}
\newfloat{listing}{tb}{lst}{}
\floatname{listing}{Listing}
\title{Listen Like a Teacher: Mitigating Whisper Hallucinations using Adaptive Layer Attention and Knowledge Distillation}
\author{
    Kumud Tripathi\equalcontrib,
    Aditya Srinivas Menon\equalcontrib,\\
    Aman Gaurav,
    Raj Prakash Gohil,
    Pankaj Wasnik
}
\affiliations{
    Sony Research India,\\
    kumud.tripathi,aditya.menon,aman.gaurav,raj.gohil,pankaj.wasnik@sony.com
}

\usepackage{bibentry}
\begin{document}

\maketitle

\begin{abstract}
The Whisper model, an open-source automatic speech recognition system, is widely adopted for its strong performance across multilingual and zero-shot settings. However, it frequently suffers from hallucination errors, especially under noisy acoustic conditions. Previous works to reduce hallucinations in Whisper-style ASR systems have primarily focused on audio preprocessing or post-processing of transcriptions to filter out erroneous content. However, modifications to the Whisper model itself remain largely unexplored to mitigate hallucinations directly. To address this challenge, we present a two-stage architecture that first enhances encoder robustness through Adaptive Layer Attention (ALA) and further suppresses hallucinations using a multi-objective knowledge distillation (KD) framework. In the first stage, ALA groups encoder layers into semantically coherent blocks via inter-layer correlation analysis. A learnable multi-head attention module then fuses these block representations, enabling the model to jointly exploit low- and high-level features for more robust encoding. In the second stage, our KD framework trains the student model on noisy audio to align its semantic and attention distributions with a teacher model processing clean inputs. Our experiments on noisy speech benchmarks show notable reductions in hallucinations and word error rates, while preserving performance on clean speech. Together, ALA and KD offer a principled strategy to improve Whisper’s reliability under real-world noisy conditions.
\end{abstract}

\section{Introduction}
Recent progress in Automatic Speech Recognition (ASR) has been fueled by Transformer-based encoder-decoder architectures, such as Whisper, which effectively capture long-range dependencies and model complex acoustic and linguistic interactions \cite{yang2024llm,tseng2024leave,radford2023robust,gulati2020conformer}. However, these models remain vulnerable to hallucinations: fluent yet semantically incorrect transcriptions. Hallucinations are particularly problematic in noisy or non‑speech segments and can seriously undermine trust in speech systems, since they often escape detection by conventional metrics like the word error rate (WER) \cite{atwany2025lost,frieske2024hallucinations}. Recent research indicates that hallucinations often arise from misaligned internal representations in both the encoder and decoder when faced with noisy input conditions \cite{atwany2025lost}. Many existing solutions focus on downstream techniques, such as pre-processing using voice activity detection \cite{bain2023whisperx}, post-processing \cite{fang2022non}, and data augmentation \cite{baranski2025investigation}, but they do not address the root cause in latent representations. To remedy this, we introduce a two‑stage framework that first strengthens encoder robustness through Adaptive Layer Attention (ALA) and mitigates hallucinations using a multi-objective knowledge distillation (KD) strategy.
% \textbf{Adaptive Layer Attention (ALA) for Encoder Robustness: }
% Transformer encoders comprise multiple layers, each capturing different levels of phonetic, lexical and semantic information. Traditional ASR models typically use only the final layer representation for decoding, potentially discarding useful intermediate features. We propose ALA, which uses inter‑layer correlation analysis to group encoder layers into semantically coherent blocks, then applies a learnable multi-head attention (MHA) mechanism \cite{vaswani2017attention} to dynamically fuse these block representations at training and inference time. By enabling segment-wise selection of the most informative layer block, ALA enhances contextual modeling in noisy speech. This adaptive fusion yields richer and more stable encoder outputs, reducing reliance on any single layer and improving noise robustness.

To enhance the robustness of encoders in noisy conditions, we introduce Adaptive Layer Attention (ALA), a dynamic fusion mechanism that utilises the hierarchical representations present within Transformer encoders. Since each transformer encoders comprise multiple layers, each capturing different levels of phonetic, lexical and semantic information \cite{layeranalysis}. Therefore, relying solely on the final layer, as traditional ASR systems do, can result in the loss of intermediate features. ALA addresses this issue by performing inter-layer correlation analysis, which groups layers into semantically coherent blocks. It then applies a learnable multi-head attention (MHA) mechanism \cite{vaswani2017attention} to dynamically fuse these block representations during both training and inference time. This enables segment-wise selection of the most informative layer block, enhancing contextual modelling under noisy conditions. By adaptively fusing information across the encoder hierarchy, ALA produces richer and more stable representations, reducing dependence on any single layer and improving overall noise robustness.

To further reduce hallucinations beyond improvements at the encoder level, we introduce a Multi-Objective Knowledge Distillation (MOKD) strategy that supervises decoder behaviour under noisy conditions. While the ALA enhances the quality of input representations, hallucinations can still occur if the decoder misinterprets noisy segments. To address this, we fine-tune the ALA-augmented encoder-decoder model using a clean-speech trained ASR model as the teacher, which provides stable representations and attention distributions. During the training of the student model on noisy inputs, we use a composite loss function that optimizes three objectives: (1) cross-entropy loss for transcription accuracy, (2) cosine similarity loss between the teacher and student representations at the final encoder and decoder layers to promote semantic and contextual alignment, and (3) Mean Squared Error (MSE) loss on the decoder's cross-attention maps to transfer the teacher's attention behavior.

Our combined framework (MOKD) first equips the encoder with noise‑aware adaptive attention, then supervises the decoder's behaviour to emulate clean speech patterns. This dual-stage design tackles hallucination at both the representation and attention levels. Related work in distillation for machine translation and speech recognition has shown that transferring attention alignment between models improves robustness and generalisation \cite{hentschel2024keep,tseng2024leave, yoon2025heuristic, nguyen2025smoothing}. Building on these insights, our approach aligns both semantic representations and attention behaviours across teacher and student models. This results in improved WER and better semantic consistency, measured via metrics such as SeMaScore \cite{sasindran2024semascore} on noisy speech datasets, while maintaining performance in clean speech. Empirical analysis demonstrates a substantial reduction in hallucination and more stable cross-attention patterns of the decoder. Our key contributions are:
\begin{enumerate}
\item We propose ALA over the Whisper encoder to enhance robustness under noisy speech conditions by incorporating low-level features from the intermediate layers.

\item We introduce a multi-loss KD strategy that combines cross-entropy, encoder/decoder cosine similarity, and decoder attention MSE to align intermediate representations with a clean teacher model.

\item We train the student model exclusively on noisy data while distilling knowledge from a clean-data teacher, enabling effective generalisation in real-world noisy scenarios.

% \item Empirical Evaluation: Extensive experiments on noisy speech benchmarks demonstrate significant improvements in word error rate and hallucination over standard finetuning and baseline KD approaches.
\end{enumerate}
%Importantly, our approach is modular, interpretable, and architecture‑agnostic, requiring only fine-tuning and no changes to decoder architecture. It offers a principled path to enhance ASR reliability in real-world noisy conditions, balancing robustness and semantic fidelity through a synergy of adaptive representation fusion and teacher-guided distillation.

\section{Related Work}
%This section provides details of literature review for Adaptive Layer Attention and Multi‑Objective Knowledge Distillation.
\textbf{Adaptive Layer Attention:}
\noindent Transformer architectures in ASR often rely on final‑layer encoder outputs, discarding intermediate representations that may carry phonetic, lexical, or semantic cues. Prior modalities, such as audio-visual ASR, have explored fusing features across layers to improve performance in noisy environments. For instance, MLCA‑AVSR applies multi-layer cross-attention fusion to combine audio and visual embeddings at different encoder depths, yielding robustness to noisy speech \cite{wang2024mlca, hentschel2024keep}.

More broadly, multi-layer feature fusion has been shown to be beneficial in vision and language modelling. While not directly targeting ASR, these studies motivate a dynamic combination of internal layer representations. Our ALA extends this principle within a single-modality ASR encoder. %We group Whisper encoder layers based on inter-layer correlation and apply segment‑wise multi‑head attention to fuse representations at inference and training time, enabling adaptive context selection per audio segment.
Although recently proposed Differential Transformer architectures introduce mechanisms to reduce attention to noise and mitigate hallucinations by amplifying relevant attention via subtraction across attention maps, these operate at the decoder head level rather than selectively fusing encoder layer representations \cite{ye2024differential}. ALA complements such efforts by enriching encoder inputs before decoding.

\noindent \textbf{Multi‑Objective Knowledge Distillation:}
\noindent Knowledge distillation (KD) has been widely used to transfer robust behaviour from one model to another, often through hidden states matching. In speech recognition, KD has improved streaming or resource-constrained ASR by aligning CTC-based teacher state alignments or token posterior outputs to a student model \cite{hentschel2024keep}.

In neural machine translation, attention alignment distillation techniques, such as “Align-to-Distill” (A2D), learn adaptive mappings between teacher and student attention heads, surpassing heuristic layer alignment and significantly boosting performance in low-resource scenarios \cite{jin2024align, inaguma2021alignment}. Our MOKD framework adapts these insights to ASR: we align encoder and decoder final-layer representations and cross-attention between a clean-speech teacher and a noisy-speech student. Our losses include encoder cosine similarity, decoder cosine similarity, and decoder cross-attention map MSE, combined with standard token-level cross-entropy. This exhaustive supervision allows the student to replicate the teacher’s robust attention behaviour and prevent hallucinations, going beyond traditional logit-only distillation. Prior works such as “Keep decoding parallel” show that intermediate-layer KD, particularly between an external language model and ASR student, can boost token accuracy in CTC-based or attention-based systems by transferring semantic LM knowledge at multiple levels \cite{jin2024align,hentschel2024keep}. Similarly, our method leverages multi-layer internal alignment, but specifically targets hallucination reduction by mirroring the attention behaviour of a clean-trained teacher.

In Distil-Whisper \cite{gandhi2023distil}, the authors used a simple word error rate (WER) heuristic to select only the highest quality pseudolabels for training. The study notes that Distil-Whisper is less prone to hallucination errors compared to the original Whisper model. Motivated by these findings, we use Distil-Whisper as one of the baseline models for this proposed approach. In combination, ALA and KD address hallucination in Whisper from two complementary angles: ALA enhances encoder representations by adaptively fusing layer blocks, improving contextual robustness under noise; KD aligns student decoder semantics and attention with that of a clean teacher. While previous literature has studied layer-wise fusion or attention alignment in isolation, our approach uniquely integrates both, yielding stronger WER reduction and hallucination mitigation in noisy ASR tasks.

\section{Proposed Methodology}
% In this work, we present a two-stage architecture aimed at improving the robustness of Whisper under noisy conditions. Our approach first enhances encoder representations using ALA referred to as Stage-1 and then mitigates hallucinations using a MOKD framework referred to as Stage-2. %This design significantly reduces the word error rate and hallucination, especially in challenging environments with speech-masked noise.

%In this section, we present the key components of our methodology: (i) Adaptive Layer Attention (ALA), which constitutes Stage 1, and (ii) a Multi-Objective Knowledge Distillation (MOKD) framework, which forms Stage 2.

This section presents the key components of our methodology: (i) Adaptive Layer Attention (ALA), which constitutes Stage-1 and enhances encoder representations, and (ii) a Multi-Objective Knowledge Distillation (MOKD) framework, which forms Stage-2 and mitigates decoder hallucinations under noisy conditions.

%This section presents details of the proposed two-stage framework. 

\begin{figure*}[!ht]
    \centering
    \includegraphics[width=0.90\linewidth, height=0.40\linewidth]{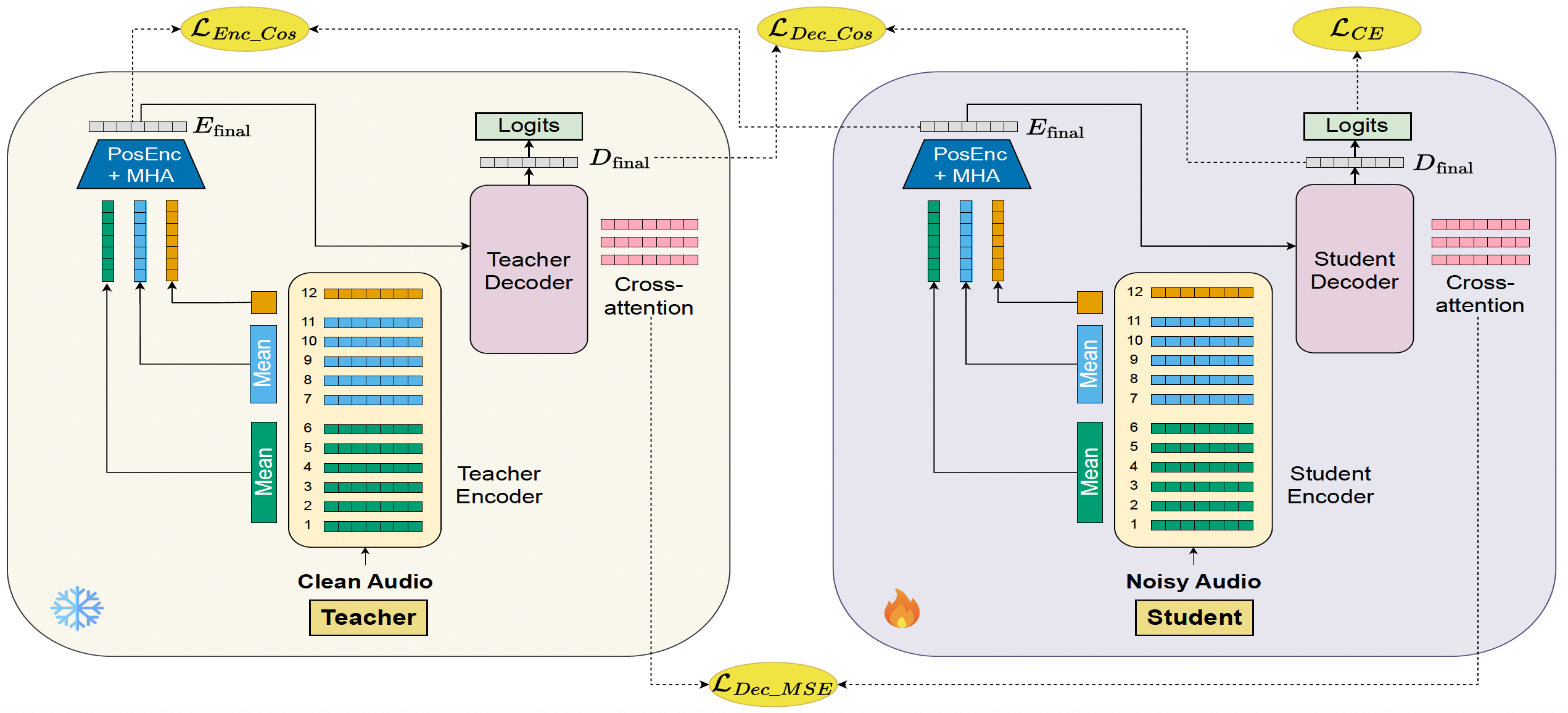}
    \caption{Block diagram of the proposed architecture combining Adaptive Layer Attention for encoder feature fusion and Multi-Objective Knowledge Distillation to reduce hallucination. $E_{final}$ and $D_{final}$ represents encoder and decoder final hidden states.}
    \label{fig:architecture}
\end{figure*}

\subsection{Adaptive Layer Attention Details}
Transformer-based ASR models like Whisper utilise deep encoder stacks, where each layer progressively abstracts features from raw audio to linguistic representations. However, under noisy conditions, certain encoder layers capture distorted or redundant signals, which degrade performance when passed directly to the decoder. To overcome this, we propose an ALA mechanism that adaptively fuses representations from structurally similar encoder layers, ensuring more robust and context-aware acoustic modelling.

\begin{figure}[!ht]
    \centering
    \includegraphics[width=0.70\linewidth, height=0.55\linewidth]{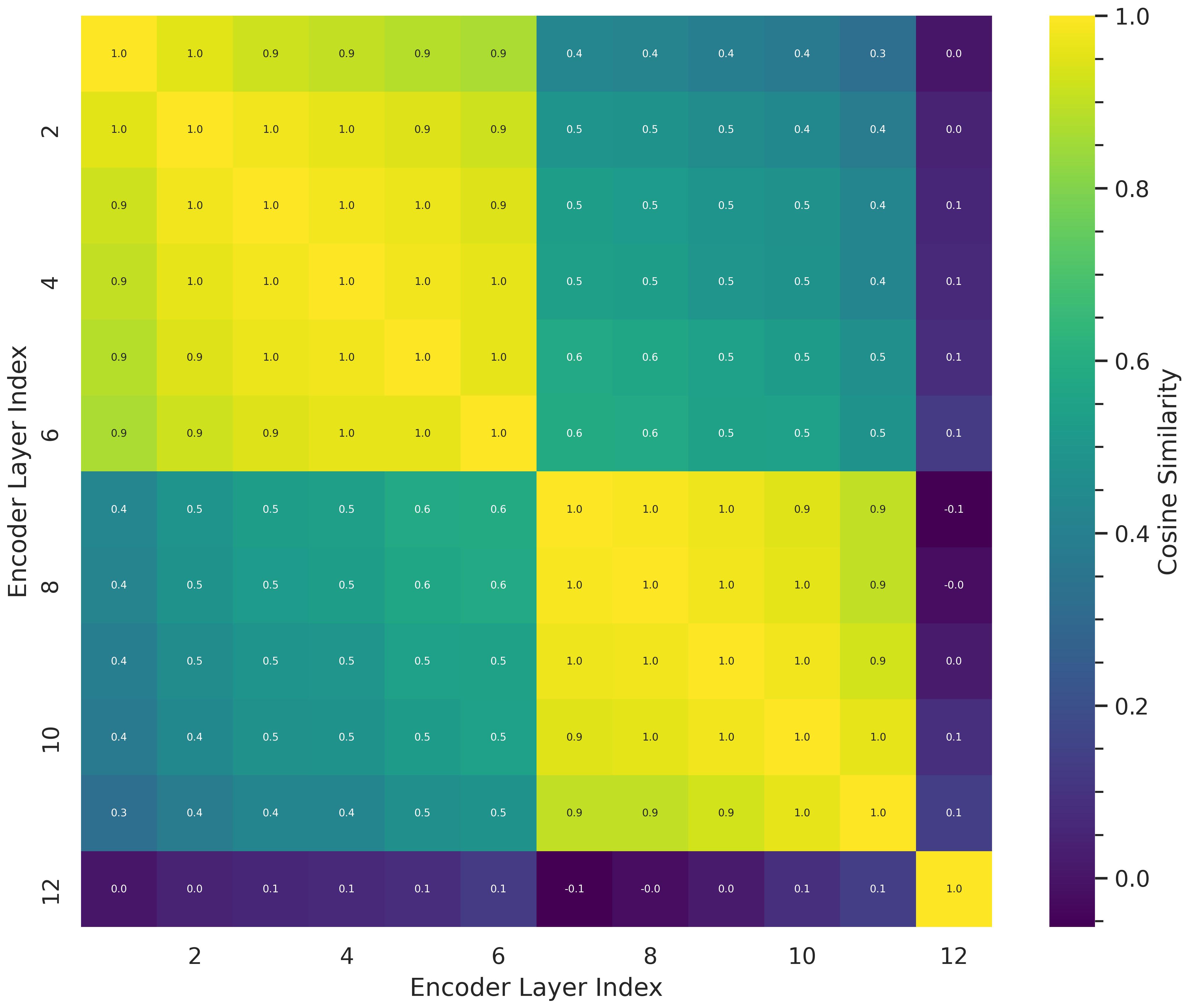}
    \caption{Heatmap for encoder inter-layer similarity on a subset of Hindi Kathbath test dataset.}
    \label{fig:encoder}
\end{figure}

\noindent{\textbf{Inter-Layer Similarity Analysis: }  }
We begin by analysing the cosine similarity between all encoder layer outputs to understand their functional roles under noise. The heatmap shown in Figure \ref{fig:encoder} reveals that layers tend to group naturally: 

\begin{itemize}
    \item Layers L1–L6 exhibit high mutual similarity, forming a block of low-level acoustic features.
    \item Layers L7–L11 form another block representing higher-level semantic abstractions.
    \item Layer L12 diverges significantly from all others, suggesting possible overfitting to noise and, notably, reflecting its specialised optimisation for decoder input.
\end{itemize}

This motivates a selective fusion strategy that avoids passing noisy or uninformative layers (e.g., L12) into the decoder, while still retaining useful abstractions. Notably, this structural pattern in encoder similarity was consistently observed across all languages evaluated in the dataset section.

\noindent{\textbf{Adaptive Fusion via Block-Wise Attention: }}
Let the encoder produce hidden states from all $L$ layers:
$$E=\{e_1,e_2,...,e_L\}$$

We first compute pairwise cosine similarity between encoder layer outputs and group layers into $K$ coherent blocks $\{B_1,B_2,...,B_K\}$, where each block contains layers with high inter-layer similarity. %Each block is assumed to capture a distinct level of abstraction (e.g., low-level acoustic vs. high-level semantic features).

\begin{enumerate}
    \item \textbf{Mean Block Representation:} 
    To derive block-level representations, we explored several strategies, including weighted summation, multi-head attention, and mean pooling \cite{gwak2025layer}. Among these, mean pooling yielded the best performance.
     Consequently, for each block $B_k$, we compute a block-level representation $r_k$ by applying the mean pooling to the hidden states of all layers in that block:
\begin{equation}
    r_k = \frac{1}{|B_k|} \sum_{l \in B_k} e_l
\end{equation}

% \[
% h_k = \frac{1}{|B_k|} \sum_{l \in B_k} e_l
% \]

This results in a set of block-wise embeddings $R = \{r_1, r_2, ..., r_K\}$.

\item \textbf{Positional Encoding:} To maintain temporal structure, we inject positional encoding (PosEnc) \cite{vaswani2017attention} into the mean block representations:
\begin{equation}
    Z=PosEnc(R)
\end{equation}
The resulting $Z$ will then be passed to the MHA.

\item \textbf{Adaptive Multi-Head Attention and Final Projection:} For each token, we use the final encoder layer’s hidden state as the query and attend over the block representations $Z$ using MHA.
% \begin{equation}
%     H = MultiHeadAttn(R)
% \end{equation}
% This step enables the model to learn cross-block dependencies, adaptively focusing on informative abstraction levels based on input context.
\begin{equation}
    h_t = \mathrm{MHA}(q_t, Z, Z)
\end{equation}
where \( q_t \) is the projected hidden state of the final encoder layer at position \( t \). The attention output is then projected, added residually to the original query, and normalised. The resulting sequence \( H = \{h_1, h_2, ..., h_T\} \) is passed to the decoder for prediction.

\end{enumerate}

\subsection{Multi-Objective Knowledge Distillation Details}
Once the encoder is enhanced with ALA, we further strengthen decoder robustness using a student-teacher KD strategy. A clean-teacher model guides a noisy-student model, aligning their encoder and decoder representations. Using a noisy teacher resulted in inferior performance, confirming the effectiveness of clean supervision for robust ASR. The distillation framework uses multiple objectives to align the student’s encoder and decoder with the clean teacher’s representations.

%Once the encoder is enhanced with ALA, we further strengthen decoder robustness using a student–teacher KD strategy. The teacher model is based on a Whisper variant trained with ALA on clean speech, while the student uses the same ALA-based model trained on noisy speech. We also experimented with a noisy teacher, but it yielded lower performance. The distillation framework incorporates multiple objectives to align both encoder and decoder representations of the student with those of the clean teacher.

%Let $(x^T, y^T)$ denote the clean teacher input-output pair and $(x^S, y^S)$ denote the corresponding noisy student input-output pair. Let $e_l^{T}$ and $e_l^{S}$ represent the hidden states at layer $l$ of the teacher and student encoder respectively. Similarly, let $d_t^{T}$ and $d_t^{S}$ represent the decoder hidden states at timestep $t$.

Let $(x^T, y^T)$ and $(x^S, y^S)$ denote the clean teacher and noisy student input-output pairs, respectively. Here, $e_t^{T}$ and $e_t^{S}$ are the encoder hidden states at timsestep $t$, while $d_t^{T}$ and $d_t^{S}$ are the decoder hidden states at timestep $t$ for the teacher and student models.

\noindent \textbf{Loss Functions: }
The student model is optimised with a combination of four objectives:

\begin{enumerate}
\item {\textbf{Encoder Cosine Similarity Loss:}} We encourage alignment between teacher and student encoder representations using cosine similarity across the last layer:
\begin{equation}
    \mathcal{L}_{Enc\_Cos} = \sum_{t=1}^{T} \left(1 - \cos \left(e^T_t, e^S_t \right) \right)
\end{equation}

\item \textbf{Decoder Cosine Similarity Loss:} To match the contextual embedding space at the decoder level, we use cosine similarity on the last layer of the decoder:
\begin{equation}
\mathcal{L}_{Dec\_Cos} = \sum_{t=1}^{T} \left(1 - \cos \left(d^T_t, d^S_t \right) \right)    
\end{equation}

%where $d^T_t$ and $d^S_t$ are the decoder hidden states of teacher and student at timestep $t$.

\item \textbf{Decoder Mean Squared Error (MSE) Loss:} We also apply MSE loss between the decoder cross-attention maps of the teacher and student:
\begin{equation}
    \mathcal{L}_{Dec\_MSE} = \sum_{t=1}^{T} \left\| d^T_t - d^S_t \right\|_2^2
\end{equation}

    \item \textbf{Cross-Entropy (CE) Loss:} The standard CE loss is used between the predicted token probabilities and the ground truth transcript:
\begin{equation}
    \mathcal{L}_{CE} = -\sum_{t=1}^{T} \log P_S(y_t)
\end{equation}

\item \textbf{Total Knowledge Distillation Loss:} The final loss combines all four components using tunable weighting coefficients:
\begin{align}
    \mathcal{L}_{\text{total}} =\ & \lambda_1 \mathcal{L}_{\text{Enc\_Cos}} + \lambda_2 \mathcal{L}_{\text{Dec\_Cos}} \nonumber \\
    & + \lambda_3 \mathcal{L}_{\text{Dec\_MSE}} + \lambda_4 \mathcal{L}_{\text{CE}}
\end{align}

% \begin{equation}
%     \mathcal{L}_{total} = \lambda_1 \mathcal{L}_{Enc\_Cos} + \lambda_2 \mathcal{L}_{Dec\_Cos} + \\ \lambda_3 \mathcal{L}_{Dec\_MSE} + \lambda_4 \mathcal{L}_{CE}
% \end{equation}

%We performed a grid search over the weighting coefficients \(\lambda_1, \lambda_2, \lambda_3, \lambda_4\) in the range 0.5 to 2.0 with increments of 0.2, and found that setting \(\lambda_1 = 0.8\) and the remaining \(\lambda\) values to 1.0 yielded the best empirical results.
We performed a grid search over the weighting coefficients \(\lambda_1, \lambda_2, \lambda_3, \lambda_4\) (ranging from 0.5 to 2.0) revealed that setting  \(\lambda_1 = 0.8\) and the others to 1.0 produced the best results.
This multi-objective KD setup ensures that the student model not only learns to mimic the teacher’s output but also captures deeper structural similarities in encoder and decoder spaces. This leads to reduced hallucination, especially in noisy speech scenarios, where decoder misalignment is more likely. 
\end{enumerate}

\begin{table*}[t]
    \caption{Comparison of Baseline-1, Baseline-2 and W-ALA models across various noise levels and clean audio for four languages. Each cell shows WER ($\downarrow$) / SeMaScore ($\uparrow$).} \label{tab:token_wer} 
\resizebox{\textwidth}{!}
{\renewcommand{\arraystretch}{1.3} 
    {\begin{tabular}{{llccccccc}} %{|c|c|c|c|c|c|c|c|c|}
    \hline
    \textbf{Language} & \textbf{Model} & \textbf{SNR -10} & \textbf{SNR -5} & \textbf{SNR 0} & \textbf{SNR 5} & \textbf{SNR 10} &  \textbf{Clean} & \textbf{Average} \\ \hline
    \textbf{Arabic} &
     Baseline-1 & \(259.59 / 0.0787\) & \(233.36 / 0.0803\) & \(235.17 / 0.0818\) & \(221.85 / 0.0820\) & \(220.07 / 0.0828\) & \(213.17 / 0.0836\) & \(230.53 / 0.0815\) \\
& Baseline-2 & \(85.80 / 0.6421\) & \(69.68 / 0.7562\) & \(61.27 / 0.8226\) & \(57.20 / 0.8551\) & \(54.34 / 0.8739\) & \(52.26 / 0.8998\) & \(63.42 / 0.8083\) \\
& W-ALA      & \(\textbf{77.51/0.6936}\) & \(\textbf{65.88/0.8008}\) & \(\textbf{58.20/0.8572}\) & \(\textbf{54.02/0.8863}\) & \(\textbf{52.06/0.8989}\) & \(\textbf{48.97/0.9118}\) & \(\textbf{59.44/0.8415}\) \\ \hline

    \textbf{French} &
 Baseline-1 & \(159.87 / 0.2912\) & \(142.05 / 0.3370\) & \(125.96 / 0.3715\) & \(116.65 / 0.3941\) & \(112.65 / 0.4049\) & \(110.44 / 0.4190\) & \(127.94 / 0.3696\) \\
& Baseline-2 & \(64.12 / 0.7583\) & \(39.29 / 0.8521\) & \(28.82 / 0.8994\) & \(24.42 / 0.9208\) & \(22.06 / 0.9312\) & \(19.76 / 0.9405\) & \(33.08 / 0.8837\) \\
& W-ALA      & \(\textbf{57.26/0.7804}\) & \(\textbf{37.15/0.8645}\) & \(\textbf{27.23/0.9076}\) & \(\textbf{23.13/0.9263}\) & \(\textbf{20.81/0.9354}\) & \(\textbf{18.91/0.9438}\) & \(\textbf{30.75/0.8930}\) \\ \hline

    \textbf{Hindi} &
     Baseline-1 & \(158.00 / 0.0897\) & \(149.53 / 0.0919\) & \(145.90 / 0.0915\) & \(132.82 / 0.0914\) & \(127.86 / 0.0911\) & \(132.38 / 0.0911\) & \(141.08 / 0.0911\) \\
& Baseline-2 & \(42.77/0.8027\) & \(26.65/0.8946\) & \(18.05/0.9365\) & \(14.95/0.9512\) & \(13.44/0.9580\) & \(12.77/0.9638\) & \(21.44/0.9178\) \\
& W-ALA      & \(\textbf{40.74/0.8257}\) & \(\textbf{24.15/0.9107}\) & \(\textbf{16.07/0.9448}\) & \(\textbf{13.33/0.9588}\) & \(\textbf{12.13/0.9636}\) & \(\textbf{11.41/0.9669}\) & \(\textbf{19.64/0.9284}\) \\  \hline
       
    \textbf{English} &
     Baseline-1 & \(40.27 / 0.8252\) & \(16.20 / 0.9237\) & \(7.11 / 0.9648\) & \(5.55 / 0.9789\) & \(4.91 / 0.9827\) & \(3.82 / 0.9853\) & \(12.97 / 0.9434\) \\
& Baseline-2 & \(39.64 / 0.8763\) & \(16.02 / 0.9434\) & \(7.21 / 0.9711\) & \(4.56 / 0.9810\) & \(3.91 / 0.9837\) & \(3.44 / 0.9851\) & \(12.46 / 0.9567\) \\
& W-ALA      & \(\textbf{29.68/0.8772}\) & \(\textbf{11.93/0.9450}\) & \(\textbf{ 5.85/0.9727}\) & \(\textbf{3.98/0.9823}\) & \(\textbf{3.45/0.9854}\) & \(\textbf{3.19/0.9866}\) & \(\textbf{9.68/0.9581}\) \\ \hline
    \end{tabular}}}
\end{table*}

\section{Experimental Setup}
\subsection{Implementation Details}
All experiments were conducted on NVIDIA H100 GPUs. %each with 80GB of memory. 
For ALA, we used 1 GPU, and MOKD was trained in a distributed fashion using 4 GPUs. Prior work \cite{atwany2025lost} highlights that smaller Whisper variants like whisper-tiny and Whisper-small exhibit higher word error rates and hallucination error rates, whereas larger models such as Whisper-medium and Whisper-large show notable improvements in performance. Based on this observation, we selected Whisper-small, referred to as W-$S$, as the base ASR model for all fine-tuning tasks in this study.
For the proposed ALA approach, we fine-tuned the W-$S$ model for 15 epochs using a learning rate of $5 \times 10^{-5}$ for the base parameters and $9 \times 10^{-5}$ for the parameters of the ALA module. The training included a warm-up phase of 5000 steps, followed by a cosine decay learning rate scheduler and 6 attention heads.
For the MOKD approach, fine-tuning was performed for 72,000 steps using a learning rate of $1 \times 10^{-5}$, with 100 warm-up steps, and a linear learning rate decay schedule.

\subsection{Dataset Details} \label{dataset}
We conducted experiments across Hindi, Arabic, French, and English to evaluate the robustness, generalizability, and effectiveness of our approach across diverse phonetic, syntactic, and noise-resilience characteristics.
%Experiments are conducted across four languages: Hindi, Arabic, French, and English. 
For Hindi ASR, we utilise the Kathbath train and test splits \cite{javed2023indicsuperb}. %for Chinese, the Aishell-1 corpus \cite{bu2017aishell}; 
Arabic and French models are trained and evaluated using the CommonVoice-15 dataset \cite{ardila2019common}, while for English ASR, we use LibriSpeech-100 for training and the test-clean split for evaluation \cite{panayotov2015librispeech}.  To replicate real-world noisy conditions and assess hallucination behaviour, we introduce noise from the Diverse Environments Multichannel Acoustic Noise Database (DEMAND) \cite{thiemann2013diverse}. This dataset includes recordings from 18 varied acoustic settings, ranging from quiet indoor environments to noisy outdoor locations, captured using a 16-channel microphone array.

Audio files exceeding 30 seconds are segmented into chunks no longer than 30 seconds. For each language, we construct noisy training datasets with uniformly distributed signal-to-noise ratios (SNRs) ranging from -8 dB to +4 dB, and corresponding test sets with SNRs from -10 dB to +10 dB. Following  \cite{zusag2024crisperwhisper}, we also include noise-only samples (without speech) labelled with empty transcripts in 1\% of the training data to further mitigate hallucinations during training.

\subsection{Comparison Methods}
% In this work, we propose a two-stage framework. The first stage introduces ALA over the Whisper encoder to enhance its robustness in noisy conditions. Building upon this, the second stage incorporates a multi-objective KD strategy to further reduce hallucinations and improve transcription quality. Our experiments consider three baselines. For the ALA-based approach, we use pre-trained Whisper-small as Baseline-1 and Whisper-small fine-tuned on noisy training dataset as Baseline-2. For evaluating the multi-objective KD model, we adopt Distill-Whisper \cite{gandhi2023distil} as Baseline-3.

%In this work, we propose a two-stage framework for mitigating hallucinations in Whisper-based ASR. 
For the first stage we apply Adaptive Layer Attention (ALA) to the Whisper encoder to improve its robustness under noisy conditions. The second stage builds on this by introducing a multi-objective Knowledge Distillation (MOKD) strategy to further reduce hallucinations and enhance transcription quality. For comparison, we evaluate three baselines: (1) the original pre-trained Whisper-small model from OpenAI, which we call Baseline-1, (2) Whisper-small fine-tuned on our noisy training set, Baseline-2, and (3) Distil-Whisper \cite{gandhi2023distil}, where the fine-tuned model from Baseline-2 serves as the teacher and a distilled student model retains all encoder layers but only two decoder layers (initialized from the teacher’s first and last decoder layers), with the encoder kept frozen. We call this Baseline-3. To ensure a fair comparison, all baselines and our proposed methods are trained and evaluated on the same noisy dataset (SNR -8 to +4 dB) described in the Dataset Details section.

\subsection{Evaluation Metrics}
%The performance of ASR transcriptions is traditionally measured using Word Error Rate (WER in \%) where lower WER is better. 
ASR performance is conventionally evaluated using Word Error Rate (WER in \%). However, as highlighted in \cite{sasindran2024semascore}, WER has notable limitations as it fails to capture semantic similarity and does not account for the varying importance of different word errors. To address this, recent works \cite{kim2021semantic,kim2021evaluating,whetten2023evaluating, sasindran2023h} have employed pretrained BERT-based models \cite{devlin2019bert} to estimate the semantic distance between the reference and hypothesis. Yet, sentence-level semantic metrics can disproportionately weight certain words, reducing their reliability. In contrast, SeMaScore \cite{sasindran2024semascore} remains effective even when WER is high, particularly under challenging conditions such as noisy speech. Accordingly, this work also evaluates hallucination in ASR outputs using SeMaScore, which ranges between 0 and 1. %Unlike WER, higher SeMaScore indicates better performance. 

\section{Results and Discussions}
In this work, we conducted different evaluations for stage-1 and stage-2 approaches. 

\begin{table}[htbp]
\caption{Latency, RTF, and peak memory usage (in GB) comparison between Baseline-2 and W-ALA models.}
\label{tab:latency_rtf_memory}
\centering
\renewcommand{\arraystretch}{1.1}
\begin{tabular}{lccc}
\hline
\textbf{Model} & \textbf{Latency (ms)} & \textbf{RTF} & \textbf{Peak VRAM} \\
\hline
Baseline-2 & 140 ± 10 & 0.021 & 1.5 \\
W-ALA      & 152 ± 11 & 0.023 & 2.6 \\
\hline
\end{tabular}
\end{table}

\begin{table*}[t]
\caption{Comparison of Baseline-2, Baseline-3 and W-MOKD models across various noise levels and clean audio for four languages. Each cell shows WER ($\downarrow$) / SeMaScore ($\uparrow$).}
\label{tab:stage2}
\resizebox{\textwidth}{!}
{\renewcommand{\arraystretch}{1.3}
\begin{tabular}{llccccccc}
\hline
\textbf{Language} & \textbf{Model} & \textbf{SNR -10} & \textbf{SNR -5} & \textbf{SNR 0} & \textbf{SNR 5} & \textbf{SNR 10} & \textbf{Clean} & \textbf{Average} \\
\hline
\multirow{2}{*}{\textbf{Arabic}} 
& Baseline-2 & \(39.64 / 0.8763\) & \(16.02 / 0.9434\) & \(7.21 / 0.9711\) & \(4.56 / 0.9810\) & \(3.91 / 0.9837\) & \(3.44 / 0.9851\) & \(12.46 / 0.9567\) \\
& Baseline-3 & \(83.53/0.6844\) & \(75.95 / 0.6038\) & \(69.77 / 0.6734\) & \(65.81 / 0.7176\) & \(63.03 / 0.7412\) & \(60.33 / 0.7628\) & \(69.74 / 0.6639\) \\
& W-MOKD      & {\(\textbf{76.32/0.7126}\)} & \(\textbf{62.85/0.8021}\) & \(\textbf{55.50/0.8719}\) & \(\textbf{51.53/0.8973}\) & \(\textbf{49.97/0.9063}\) & \(\textbf{48.21/0.9202}\) & \(\textbf{57.40/0.8557}\) \\
\hline
\multirow{2}{*}{\textbf{French}}
& Baseline-2 & \(64.12 / 0.7583\) & \(39.29 / 0.8521\) & \(28.82 / 0.8994\) & \(24.42 / 0.9208\) & \(22.06 / 0.9312\) & \(19.76 / 0.9405\) & \(33.08 / 0.8837\) \\
& Baseline-3 & \(63.60 / 0.6988\) & \(47.12 / 0.7877\) & \(38.98 / 0.8301\) & \(34.02 / 0.8515\) & \(31.64 / 0.8616\) & \(28.23 / 0.8796\) & \(40.60 / 0.8181\) \\
& W-MOKD      & \(\textbf{51.41/0.7690}\) & \(\textbf{34.42 / 0.8767}\) & \(\textbf{25.44/0.9152}\) & \(\textbf{21.40/0.9318}\) & \(\textbf{19.61/0.9393}\) & \(\textbf{17.79/0.9468}\) & \(\textbf{28.18/0.8965}\) \\
\hline
\multirow{2}{*}{\textbf{Hindi}}
& Baseline-2 & \(42.77/0.8027\) & \(26.65/0.8946\) & \(18.05/0.9365\) & \(14.95/0.9512\) & \(13.44/0.9580\) & \(12.77/0.9638\) & \(21.44/0.9178\) \\
& Baseline-3 & \(56.16 / 0.6950\) & \(54.72 / 0.7412\) & \(48.20 / 0.7755\) & \(46.53 / 0.7757\) & \(46.28 / 0.7749\) & \(47.43 / 0.7602\) & \(51.39 / 0.7544\) \\
& W-MOKD      & \(\textbf{38.13/0.8455}\) & \(\textbf{22.67/0.9257}\) & \(\textbf{14.83/0.9580}\) & \(\textbf{12.97/0.9697}\) & \(\textbf{11.86/0.9644}\) & \(\textbf{11.23/0.9675}\) & \(\textbf{18.61/0.9432}\) \\
\hline
\multirow{2}{*}{\textbf{English}}
& Baseline-2 & \(39.64 / 0.8763\) & \(16.02 / 0.9434\) & \(7.21 / 0.9711\) & \(4.56 / 0.9810\) & \(3.91 / 0.9837\) & \(3.44 / 0.9851\) & \(12.46 / 0.9567\) \\
& Baseline-3 & \(69.75 / 0.7634\) & \(55.79 / 0.8327\) & \(44.69 / 0.8660\) & \(45.36 / 0.8892\) & \(42.47 / 0.8767\) & \(44.11 / 0.9029\) & \(50.36 / 0.8592\) \\
& W-MOKD      & \(\textbf{26.43/0.9046}\) & \(\textbf{9.41/0.9726}\) & \(\textbf{5.72/0.9842}\) & \(\textbf{3.11/0.9853}\) & \(\textbf{3.49/0.9832}\) & \(\textbf{3.18/0.9836}\) & \(\textbf{8.56/0.9690}\) \\
\hline
\end{tabular}}
\end{table*}
\subsection{Stage-1 Evaluation}
Table \ref{tab:token_wer} presents the first stage of evaluation, where we examine the effectiveness of the ALA mechanism across four languages: Arabic, French, Hindi, and English, under a range of noise conditions, from –10 dB SNR to clean speech.
Compared to both Baseline-1 and Baseline-2, the Whisper finetuned with ALA, referred to as the W-ALA model, consistently achieves lower WER and higher SeMaScore, demonstrating enhanced robustness and semantic preservation in noisy environments. Notably, proposed W-ALA offers significant improvements at low SNRs (e.g., -10 dB, 5dB and 0 dB), where conventional baselines degrade sharply. On average, W-ALA reduces WER by a substantial margin while simultaneously increasing cosine similarity, indicating more reliable and confident predictions. These results validate the ALA module’s ability to dynamically prioritise acoustically informative encoder layers, improving recognition performance without compromising semantic correctness.

We benchmarked the inference efficiency of the W-ALA model against Baseline-2 on the LibriSpeech English test set using a batch size of 1. As shown in Table~\ref{tab:latency_rtf_memory}, W-ALA adds only minimal overhead: latency increases by 8\%, real-time factor (RTF) by 9\%. Peak VRAM usage rises by approximately 1 GB, remaining well within the capacity of modern accelerators. Despite introducing just 0.98\% more parameters, W-ALA delivers significant gains in robustness and accuracy (see Table \ref{tab:token_wer}) over Baseline-2. These results highlight that W-ALA achieves improved performance with negligible impact on runtime, memory usage and additional parameters, making it a practical and scalable solution for real-world ASR applications.
\subsection{Stage-2 Evaluation}
Table \ref{tab:stage2} presents a comprehensive evaluation of the proposed W-MOKD model across four languages: Arabic, French, Hindi, and English, under a range of noisy (SNR-10 to 10 dB) and clean conditions. The results demonstrate that W-MOKD consistently outperforms Baseline-2 and the strong Distil-Whisper baseline (Baseline-3) in both WER and SeMaScore across all languages and SNR levels. Notably, the gains are more significant under severe noise (e.g., at -10 dB). Even under clean conditions, W-MOKD yields improvements in semantic accuracy. Moreover, W-MOKD outperforms the W-ALA model. These results validate the effectiveness of our two-stage architecture in enhancing robustness and reducing hallucinations across diverse linguistic and acoustic settings, particularly under challenging noise conditions.

%Moreover, W-MOKD outperforms the W-ALA model, indicating that integrating Adaptive Layer Attention with multi-objective knowledge distillation further enhances robustness. These results substantiate the effectiveness of the proposed two-stage architecture in suppressing hallucinations and delivering reliable ASR performance across diverse languages and varying noise scenarios.

%The Table \ref{tab:stage2} presents evaluation results from Stage-2 (MOKD) across multiple languages under varying noise conditions (SNR from –10 dB to clean). The comparison is made between Baseline-3 (Distill-Whisper) and the proposed W-MOKD model. Across 4 languages the W-MOKD consistently outperforms Baseline-3, particularly under high-noise scenarios (e.g., –10 dB and –5 dB), where hallucinations are most prominent. 
%The W-MOKD is even better then the W-ALA model.
%For example, in Hindi, WER drops from 56.16\% (Baseline-3) to 39.93\% at –10 dB, while the SeMaScore improves from 0.6944 to 0.8536. Similar trends are observed in French, where the WER at –5 dB reduces from 47.12\% to 34.42\%, with corresponding SeMaScore improvements. Arabic also shows substantial improvement, with an average WER drop from 69.74\% to 57.40\% and SeMaScore rising from 0.6639 to 0.8557.

The average WER and SeMaScore across noise levels show clear gains, highlighting that the proposed MOKD framework effectively suppresses hallucinations and enhances semantic accuracy, particularly in challenging acoustic conditions. 
Overall, these results validate the effectiveness of combining adaptive encoder attention with attention- and representation-level distillation, resulting in more noise-robust and semantically accurate ASR across multiple languages.

\subsection{Analysing Encoder Block Robustness}
%The Figure \ref{fig:robust} provides a comprehensive analysis of  the layer attention behavior in the context of noisy speech processing, emphasizing the robustness of different encoder blocks. The heatmap (Figure \ref{fig:fig1}) visualizes the dynamic attention weights assigned to each encoder block over time for a representative noisy audio sample. It reveals that Block 0 consistently receives higher attention across a majority of frames, particularly in segments likely affected by noise or weak signal quality. This suggests that Block 0 captures more stable representations during inference. In contrast, Blocks 1 and 2 show infrequent and comparatively lower attention contributions, indicating that the model selectively relies less on deeper or higher-level features under noisy conditions. Complementing this, the bar chart (Figure \ref{fig:fig2}) presents the average attention weights across the entire dataset, highlighting a clear preference for Block 0 with an average weight of 0.586, while Blocks 1 and 2 receive significantly lower and nearly equal weights. This consistent trend across both individual samples and the full dataset demonstrates the robustness and reliability of Block 0 in noisy acoustic environments, validating the effectiveness of the ALA mechanism in prioritizing stable encoder features.

\begin{figure*}[ht]
    \centering
    \begin{subfigure}[t]{0.45\textwidth}
        \centering
        \includegraphics[width=0.88\linewidth]{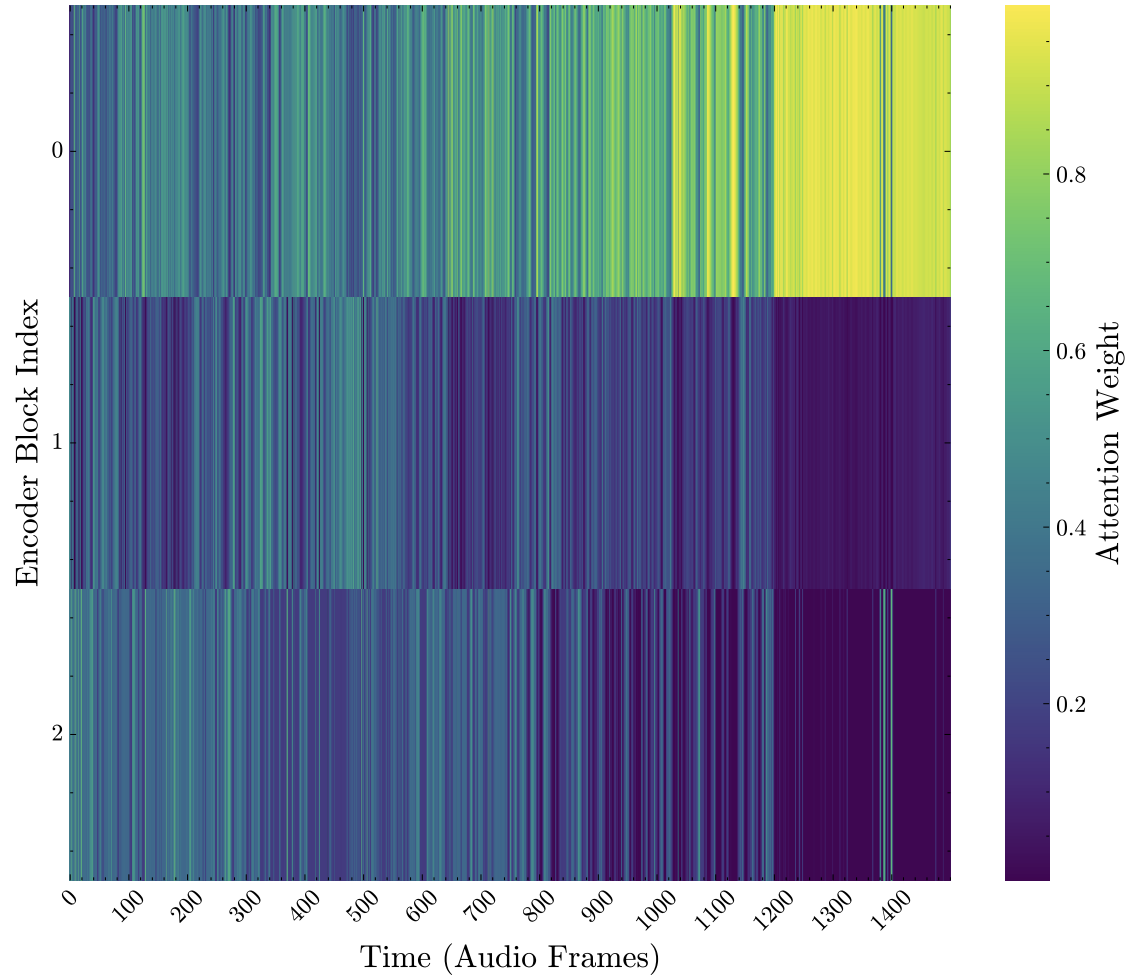}
        \caption{Block attention across time for a sample}
        \label{fig:fig1}
    \end{subfigure}
    \hfill
    \begin{subfigure}[t]{0.45\textwidth}
        \centering
        \includegraphics[width=0.88\linewidth]{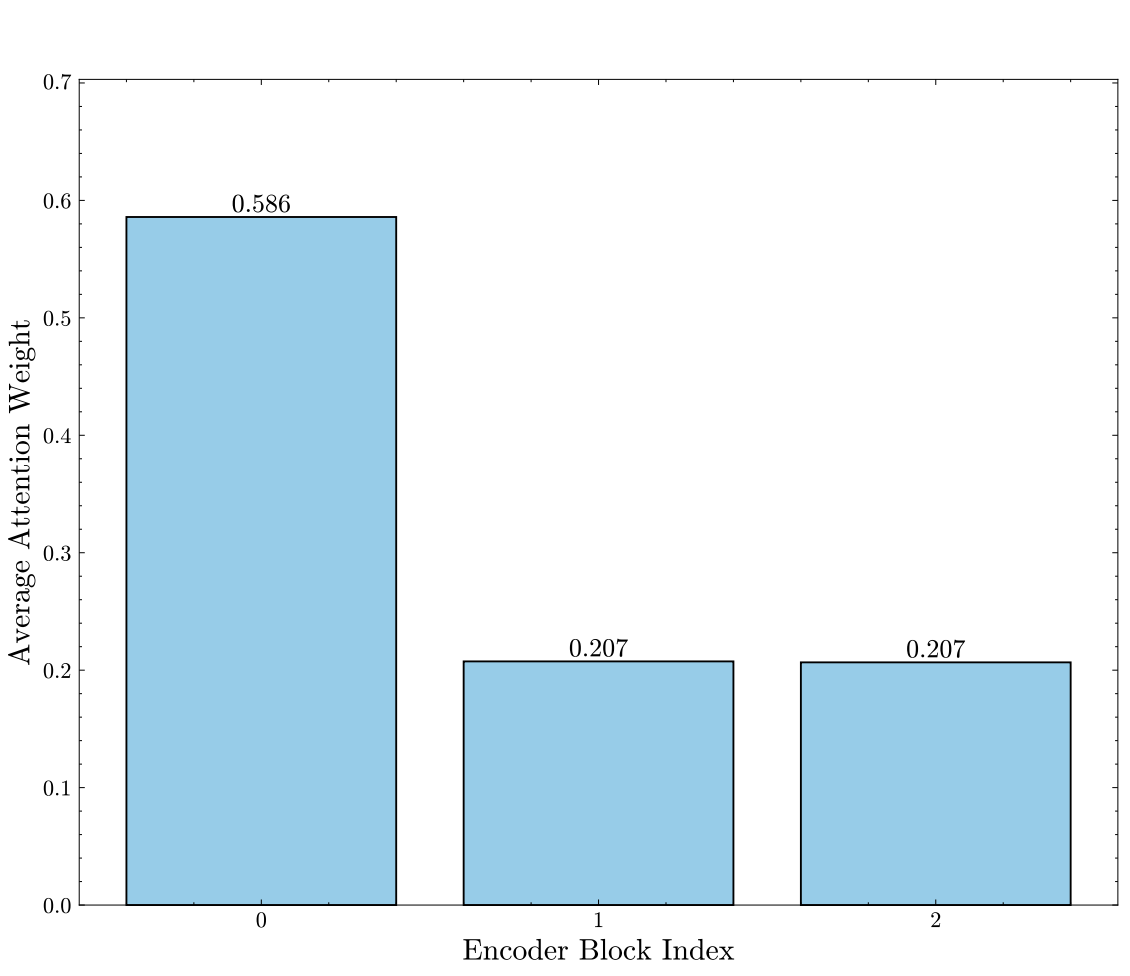}
        \caption{Dataset-wide average attention weights}
        \label{fig:fig2}
    \end{subfigure}
    \caption{Layer-wise attention behaviour for noisy speech input.}
    \label{fig:robust}
\end{figure*}

Figure \ref{fig:robust} presents an analysis of encoder block attention behaviour under noisy conditions, highlighting the robustness of different blocks in the ALA mechanism. The heatmap (Figure \ref{fig:fig1}) shows frame-wise attention distribution for a noisy audio sample, where Block 0 consistently receives the highest attention, especially during segments likely impacted by noise. This indicates that Block 0 captures more stable and noise-resilient features during inference. In contrast, Blocks 1 and 2 are less frequently attended, suggesting the model relies less on deeper or more abstract representations in noisy scenarios. The accompanying bar chart (Figure \ref{fig:fig2}) shows average attention weights across the entire dataset, reinforcing this pattern: Block 0 dominates with an average weight of 0.586, while Blocks 1 and 2 receive significantly lower and nearly equal attention. Together, these results demonstrate a consistent model preference for Block 0 in noisy settings. This validates the effectiveness of Adaptive Layer Attention (ALA) in dynamically emphasising robust encoder features for improved ASR performance.

\subsection{Stage-1 Ablation Study}
In Table \ref{tab:loss_function_comparison}, the Stage-1 ablation study investigates various strategies for fusing encoder representations in Whisper using ALA, conducted on the Hindi dataset at –10 dB SNR and clean. The first method, a weighted sum across all encoder layers, performs the worst (WER: 75.85/29.56, SeMaScore: 0.4822/0.7521) on -10 dB and clean, likely due to uniformly treating layers without capturing their functional diversity. Applying MHA over all encoder layers while keeping them frozen yields a slightly better WER (52.73/15.62), indicating frozen layers cannot adapt to noise. When layers are trainable during MHA, performance improves significantly (WER: 45.12/14.87, SeMaScore: 0.6902/0.9290), demonstrating that encoder layers begin learning noise-robust features.

We further group encoder layers into three blocks based on inter-layer similarity and experiment with MHA on first (1,7,12), last (6,11,12), and middle (3,9,12) layers of each block. These configurations yield improved WERs and better SeMaScores. The best results (WER: 40.74/11.41, SeMaScore: 0.8257/0.9669) come from applying MHA over mean representations from each block, highlighting the benefit of block-level context aggregation for robust and hallucination-resistant ASR.
\newcolumntype{Y}{>{\centering\arraybackslash}X}
\begin{table}[htbp]
  \caption{Comparison of encoder-fusion strategies under noisy (-10 dB) and clean conditions on the Hindi test set.}
  \label{tab:loss_function_comparison}
  \renewcommand{\arraystretch}{1.1}
  \begin{tabularx}{\linewidth}{lYY}
    \hline
    \textbf{Fusion Method} & \textbf{SNR -10} & \textbf{Clean} \\
    \hline
    Weighted Sum   & \(75.85/0.4822\) & \(29.56/0.7521\) \\
    MHA all frozen & \(52.73/0.5454\) & \(15.62/0.8928\) \\
    MHA all        & \(45.12/0.6902\) & \(14.87/0.9290\) \\
    MHA 1,7,12     & \(42.23/0.7032\) & \(14.20/0.9352\) \\
    MHA 6,11,12    & \(41.91/0.7105\) & \(12.16/0.9480\) \\
    MHA 3,9,12     & \(41.53/0.7135\) & \(12.14/0.9492\) \\
    MHA Mean       & \(\mathbf{40.74/0.8257}\) & \(\mathbf{11.41/0.9669}\) \\
    \hline
  \end{tabularx}
\end{table}

\subsection{Stage-2 Ablation Study}
Table \ref{tab:kd_wer_sema_comparison} presents an ablation study assessing the impact of different KD loss functions under both -10 dB and clean conditions. Here, CosAll applies cosine similarity loss across all encoder and decoder hidden states, while CosFin restricts it to only the final layers. Similarly, Optimal Transport OTFin, IRLEFin, and KLFin apply OT \cite{mao2025weighted}, IRLE \cite{higuchi2021noise}, and KL divergence losses to final encoder and decoder representations. MSE\(_{\text{decCA}}\) denotes mean squared error loss between decoder cross-attention maps, and MSE\(_{\text{decSA}}\) is between decoder self-attention maps, and MSE\(_{\text{enc}}\) represents loss between encoder attention maps.

Among the methods, CosFin offers a strong trade-off, showing that deep-layer alignment is more semantically meaningful, performing better alignment over CosAll. KLFin provide marginal gains, while distance-based losses like IRLE and OT significantly degrade performance. Adding only MSE\(_{\text{decCA}}\) offers the best improvement across all attention losses. The best results emerge when combining CosFin and MSE\(_{\text{decCA}}\) with ALA. %achieving WER 39.93\% and SeMaScore 0.8356.

\begin{table}[t]
  \caption{Comparison of KD loss functions under noisy
           (-10 dB) and clean conditions.}
  \label{tab:kd_wer_sema_comparison}
  \footnotesize
    \renewcommand{\arraystretch}{1}
  \setlength{\tabcolsep}{4pt}     % tighten horizontal padding
  \begin{tabularx}{\linewidth}{lcc}
    \toprule
    \textbf{KD Loss Function}                        &
    \textbf{SNR -10} &
    \textbf{Clean} \\
    \midrule
    CE (no KD)                             & \(42.77 / 0.8027\) & \(12.77 / 0.9638\) \\
    \midrule
    +\,KL\(_{\text{logits}}\)              & \(46.48 / 0.7331\) & \(14.61 / 0.9413\) \\
    \midrule
    +\,CosAll                              & \(45.61 / 0.8071\) & \(14.61 / 0.9592\) \\
    \textbf{+}\,\textbf{CosFin}                              & \(\textbf{42.97} / \textbf{0.8246}\) & \(\textbf{12.37} / \textbf{0.9622}\) \\
    +\,OTFin                               & \(46.53 / 0.8019\) & \(13.56 / 0.9614\) \\
    +\,IRLEFin                             & \(61.14 / 0.6102\) & \(39.36 / 0.7333\) \\
    +\,KLFin                               & \(43.88 / 0.8193\) & \(13.15 / 0.9625\) \\
    \midrule
    +\,CosFin+MSE\(_{\text{decCA}}\)                      & \(42.46 / 0.7207\) & \(12.52 / 0.9486\) \\
    +\,CosFin+MSE\(_{\text{decCA}}\)+MSE\(_{\text{decSA}}\)                   & \(42.82 / 0.7192\) & \(12.97 / 0.9373\) \\
    +\,CosFin+MSE\(_{\text{decCA}}+\)MSE\(_{\text{enc}}\)  & \(42.73 / 0.7137\) & \(12.94 / 0.9345\) \\
    \textbf{+}\,\textbf{CosFin+MSE}\(_{\text{\textbf{decCA}}}\)\textbf{+ALA }                 & \(\textbf{38.13} / \textbf{0.8455}\) & \(\textbf{11.23} / \textbf{0.9675}\) \\
    \bottomrule
  \end{tabularx}
\end{table}

\section{Conclusions}
In this work, we proposed a two-stage framework to improve the robustness and semantic fidelity of Whisper model under noisy conditions. In Stage-1, we introduced an adaptive layer attention mechanism that dynamically fuses encoder representations based on inter-layer similarity. This approach enables the model to emphasise the most informative and noise-resilient encoder blocks, effectively suppressing irrelevant activations that often lead to hallucinations. In Stage-2, we enhanced this with a multi-objective knowledge distillation framework that aligns the student model’s encoder representations, decoder semantics, and cross-attention maps with those of a clean teacher. Our experiments across multiple languages and varying noise levels demonstrate substantial improvements in both word error rate and SeMaScore, particularly under challenging low-SNR conditions. Overall, our framework offers a principled approach to mitigate hallucinations and improve Whisper reliability in real-world noisy environments. Future work can explore cross-lingual generalisation and apply our proposed approach to other transformer-based speech models.
\bibliography{aaai2026}

\end{document}